\documentclass{article}
\usepackage{spconf,amsmath,epsfig}
\usepackage{graphicx}
\usepackage[colorlinks,linkcolor=blue]{hyperref}
\usepackage{amssymb}
\usepackage{indentfirst}
\usepackage[T1]{fontenc}
\usepackage{multirow}

\let\OLDthebibliography\thebibliography
\renewcommand\thebibliography[1]{
  \OLDthebibliography{#1}
  \setlength{\parskip}{0pt}
  \setlength{\itemsep}{0pt plus 0.3ex}
}

\begin{document}\sloppy

\def\x{{\mathbf x}}
\def\L{{\cal L}}

\title{PSSTRNet: Progressive Segmentation-guided Scene Text Removal Network}
%
\name{Guangtao Lyu, Anna Zhu\footnotemark{*}}
\address{School of Computer Science and Artificial Intelligence, Wuhan University of Technology, China\\
annazhu@whut.edu.cn}

\maketitle
\renewcommand{\thefootnote}{\fnsymbol{footnote}} 
\footnotetext[1]{Corresponding author} 

\begin{abstract}
Scene text removal (STR) is a challenging task due to the complex text fonts, colors, sizes, and background textures in scene images. However, most previous methods learn both text location and background inpainting implicitly within a single network, which weakens the text localization mechanism and makes a lossy background. To tackle these problems, we propose a simple Progressive Segmentation-guided Scene Text Removal Network(PSSTRNet) to remove the text in the image iteratively. It contains two decoder branches, a text segmentation branch, and a text removal branch, with a shared encoder. The text segmentation branch generates text mask maps as the guidance for the regional removal branch. In each iteration, the original image, previous text removal result, and text mask are input to the network to extract the rest part of the text segments and cleaner text removal result. To get a more accurate text mask map, an update module is developed to merge the mask map in the current and previous stages. The final text removal result is obtained by adaptive fusion of results from all previous stages. A sufficient number of experiments and ablation studies conducted on the real and synthetic public datasets demonstrate our proposed method achieves state-of-the-art performance. The source code of our work is available at: \href{https://github.com/GuangtaoLyu/PSSTRNet}{https://github.com/GuangtaoLyu/PSSTRNet.}
\end{abstract}

\begin{figure*}
  \includegraphics[width=\textwidth,height=0.35\textheight]{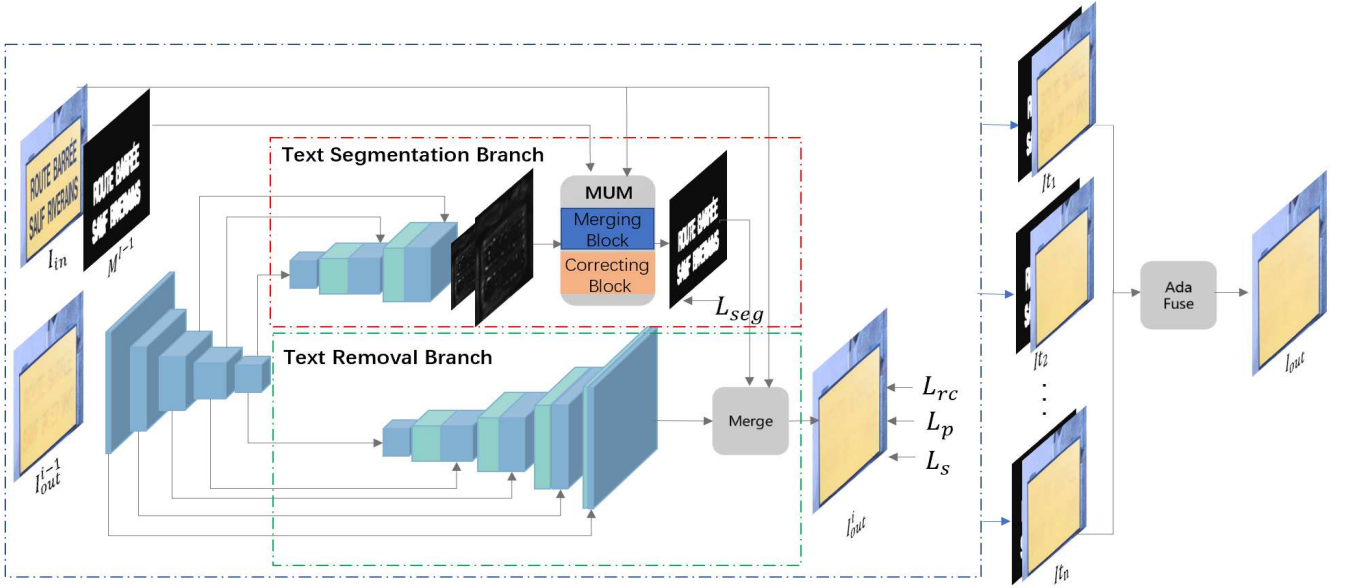}
  \caption{ The overall architecture of PSSTRNet. TSB is the text segmentation branch that outputs a mask with $\frac{1}{4}$ size of input text image $I_{in}$. The mask is further corrected in the Mask-Update block. TRB is the text removal branch, and it outputs the temporary text removed result $I_{temp}$. Then, $I_{temp}$ and $I_{in}$ are merged in the Merge block to get correct text removed result $I_{out}^i$ in current iteration. The final result is the adaptive fusion of results from all iterations.}
  \label{model}
\end{figure*}

\section{Introduction}
\label{sec:intro}

Scene text contains quite a lot of sensitive and private information. To prevent the private information in images from being used illegally, scene text removal(STR) is proposed to address this issue.

The well-known Pix2Pix\cite{2017Image} which used patch-GAN for image translation can be applied to the STR task. So, Scene Text Eraser(STE)\cite{2017Scene} adopted its idea and used a single-scaled sliding window to remove the text in each patch independently. This method processed STR locally without considering the global context information. EnsNet\cite{2018EnsNet} developed several designed loss functions and a lateral connection to further enhance the STR performance. However, these single stage-based STR methods may modify non-text pixels in images and result in excessive or inadequate inpainting results. MTRNet\cite{2019MTRNet} employed conditional GAN and used the text segmentation results for inpainting region guidance. EraseNet\cite{2020EraseNet} used a text detection branch to locate text regions and remove text from coarse to fine. However, the STR performance of those methods relay heavily on only one-shot text segmentation results. PERT\cite{wang2021pert} performed multi-stage text erasure in a progressive way\cite{2019Progressive} with explicit text region guidance. However, it could not get more accurate text regions in iteration stages, and the network was difficult to train.

In this paper, we propose a Progressive Segmentation-guided Scene Text Removal Network (PSSTRNet) with very low computation costs. It is built on a very simple and small network, which has one feature-sharing encoder and two decoders to generate text segmentation and removal results individually. However, we find that single forward computing generates very coarse STR results. So, we input the text removal image to the network again to yield refined results progressively. A Mask Update module is added in the text segmentation branch for generating more precise text segmentation results. Additionally, we design an adaptive fusion method to make full use of the results of different iterations.

We conducted sufficient experiments on two datasets: SCUT-EnsText\cite{2020EraseNet} and SCUT-syn\cite{2018EnsNet}. Both the qualitative and quantitative results indicate that PSSTRNet can outperform previous state-of-the-art STR methods.

We summarize the contributions of this work as follows:
\begin{itemize}
\item We propose a novel STR network termed PSSTRNet. It decomposes the challenging STR task into two simple subtasks and processes text segmentation and background inpainting progressively.
\item We design a Mask Update module and an adaptive fusion strategy to make full use of results from different iterations.
\item Our proposed PSSTRNet is light-weighted and achieves SOTA quantitative and qualitative results on public synthetic and real scene datasets.
\end{itemize}

\section{Proposed Method}

\subsection{Overall pipeline}

As shown in Fig.\ref{model}, the pipeline of our model consists of two branches: the text segmentation branch and the text removal branch. They share a lightweight encoder with 5 residual convolutional layers. PSSTRNet implements text segmentation and erasing process on the previous results iteratively, and merges all the results in each iteration as final output adaptively.

\subsection{Text Segmentation Branch}\label{sect1}
The text segmentation branch contains a Text Region Positioning module (i.e., TRPM), an upsampling process, and a Mask Updating module(i.e. MUM). To reduce the computational cost, TRPM is designed to locate the unremoved text regions in text removal images from the output of the previous iteration. It outputs the expansion text mask with 1/4 size of the original image. Then, this mask goes through the upsampling process by two bilinear interpolations to get the mask owning the same size as the origin image. The size recovered expansion text mask is denoted as $M_{temp}^{i}$ ($M_{temp}^{i}\in[0,1]$, 1 for text region and 0 for non-text region). With the input of the previous text mask $M^{i-1}$ and $M_{temp}^{i}$, MUM updates $M^{i-1}$ and outputs the final text mask map($M^i$) in $i_{th}$ iteration. It includes a Merging block and a Correcting block. The Merging block merges $M_{temp}^{i}$ and $M^{i-1}$ through Eq.\eqref{eq1} to get a more complete text mask map $M_{comp}^{i}$.
\begin{equation}
 M_{comp}^{i}= max(M_{temp}^{i},M^{i-1})
  \label{eq1}
\end{equation}
In Correcting block\cite{mei2021camouflaged}, $M_{comp}^{i}$ is first multiplied with the origin text image $I_{in}$, to generate the text-attentive features $I_t$ and the background-attentive features $I_b$, respectively. Then, we feed these two types of features into two parallel context exploration (CE) blocks to perform contextual reasoning for discovering the false-positive distractions $I_{fp}$ and the false-negative distractions $I_{fn}$, respectively. The CE block consists of four dilation convolutions with different dilation rates of 1, 2, 3, 5. The outputs of all the four dilation convolutions are concatenated and then fused via a 1$\times$1 convolution. Using such a design, the CE block gets the capability of exploring abundant context over a wide range of scales and thus can be used for context reasoning.

After context reasoning, we can correct the mask in the following way:
\begin{equation}
\begin{split}
& I_{in} = {\rm NR}(I_{in} - \alpha I_{fp}),\\
& I_{in} = {\rm NR}(I_{in} + \beta I_{fn})), \\
& M^i = \sigma (I_{in}).
  \label{eq2}
\end{split}
\end{equation}

\noindent where $\alpha$ and $\beta$ are the learnable parameters that are initialized as 1. NR is batch normalization and ReLU operation. $\sigma$ is the sigmoid function.

We use the element-wise subtraction operation to restrain the ambiguous backgrounds (i.e., false-positive distractions) and the element-wise addition operation to enhance the missing text regions (i.e., false-negative distractions). Then, we apply a sigmoid function to get a more precise text mask map $M^i$.

\subsection{Text Removal Branch}\label{sect2}

Similarly, the text removal branch and the shared encoder are built on a simplified residual U-Net structure. The encoder has five convolution layers with kernel size k$\times$k (k=7,5,3,3,3 in each layer in order). Each layer contains a batch normalization, a ReLU, and a residual block after convolution operation. With inputting the previous result $I_{out}^{i-1}$ in $i_{th}$ iteration, the output is defined as $I_{out}^i$.

The goal of STR is to remove the text areas while keeping the background areas unchanged. So, we merge the text regions of $I_{out}^i$, and non-text regions of $I_{in}$ as the final output $I_{out}^i$ of $i_{th}$ iteration as in Eq.\eqref{eq3}.
\begin{equation}
 I_{out}^i = I_{in}^i*(1-M^i) + I_{out} * M^i,
  \label{eq3}
\end{equation}
\noindent Where $I_{in}$ is the original text image.

Finally, after a specific number of iterations, the text regions can be extracted more accurately and the text erased cleaner. However, since the process of mapping RGB images to the latent features and mapping them back to the RGB space occurs in each iteration, it results in information distortion in every recurrence. To solve these problems, we merge the intermediate iteration outputs in an adaptive merging way as formulated in Eq.\eqref{eq4}. The final output is $I_{out}$.

\begin{equation}
\begin{split}
	& M^{'} = \frac{\sum_{1}^{n} M^i}{n}\\
	& I_{out}^{'} = \frac{\sum_{1}^{n} I_{out}^i * M^i}{n}\\
	& I_{out}^{'} = (I_{out}^{'} + \epsilon)/(M^{'} + \epsilon)\\
	& I_{out} = I_{in}^{'}* (1-M^{'}) + I_{out}*M^{'}\\
  \label{eq4}
\end{split}
\end{equation}
where $\epsilon$ is a smoothing factor and is set to be $1e^{-8}$

\begin{figure*}
  \includegraphics[width=\linewidth,height=0.35\linewidth]{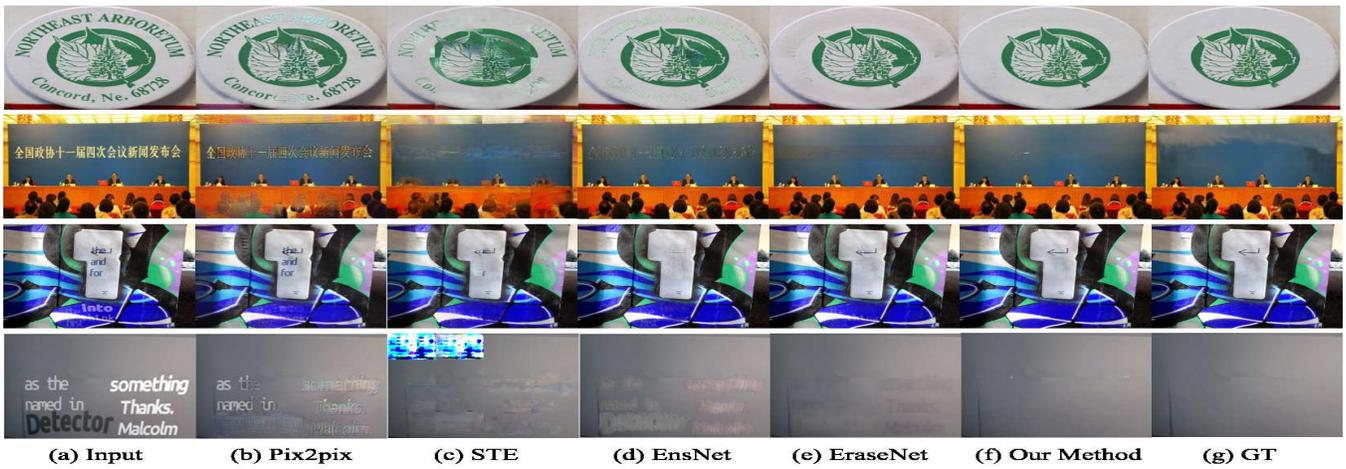}
  \caption{Comparison results with other SOTA methods on SCUT-EnsText and SCUT-Syn datasets.}
  \label{SOTA}
\end{figure*}

\begin{figure*}
  \includegraphics[width=\linewidth]{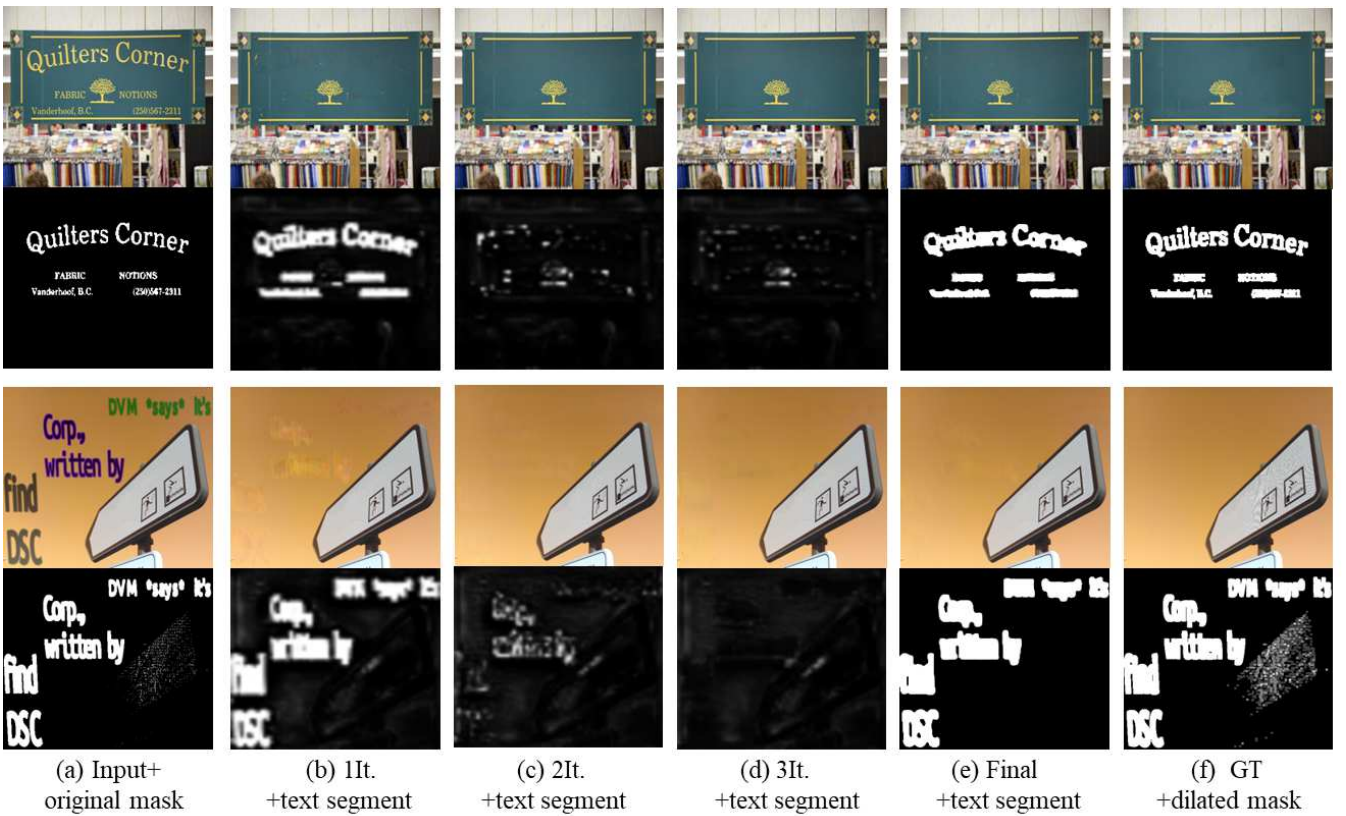}
  \caption{The results of different iterations on SCUT-EnsText and SCUT-Syn datasets. It$_i$ is the $M_{temp}$ of $i_{th}$ iteration. Final represents the final STR results and final fused mask.}
  \label{Diff_stage}
\end{figure*}

\begin{figure*}
  \vspace{-0.7cm}
  \includegraphics[width=\linewidth,height=0.35\linewidth]{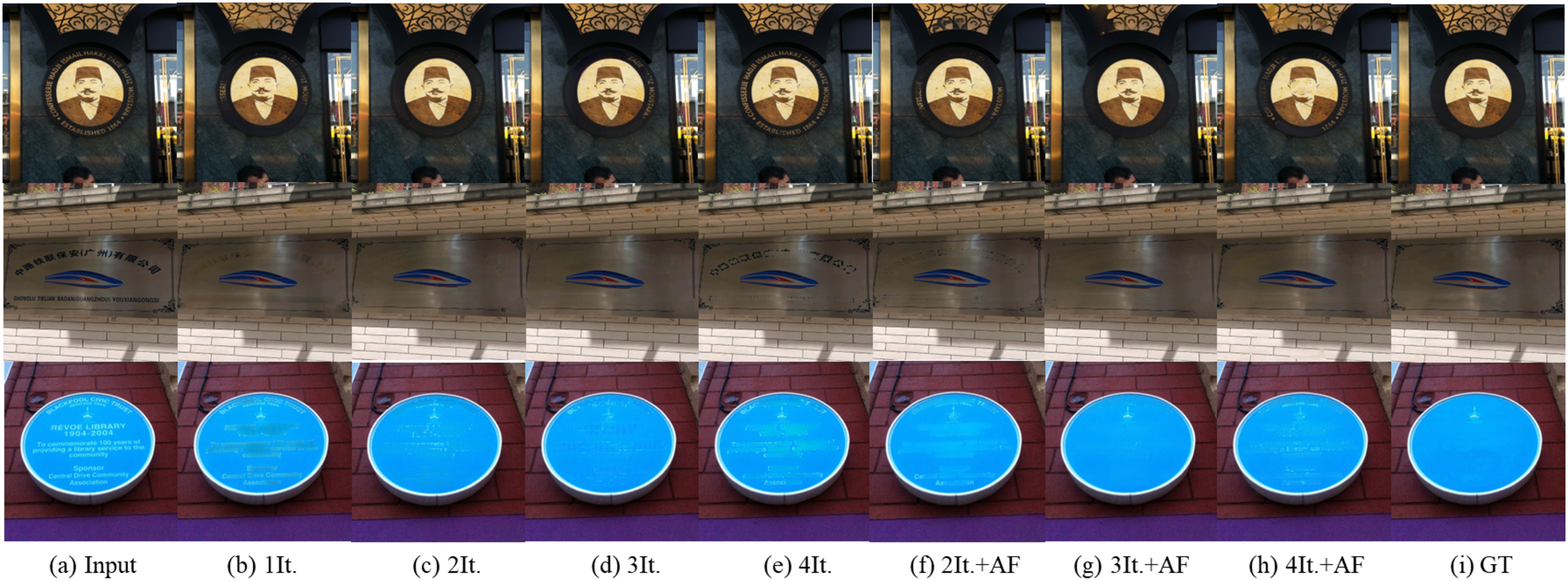}
  \caption{The results of ablation study on SCUT-EnsText dataset.}
  \label{Ablation}
\end{figure*}

\subsection{Loss Functions}\label{sect3}

We introduce several loss functions for PSSTRNet learning, including region content loss $L_{rc}$, perceptual loss $L_{p}$, style loss $L_{s}$ and segmentation loss $L_{seg}$. Given the origin text image $I_{in}$, text-removed ground truth (gt) $I_{gt}$ and the binary text gt mask $M_{gt}$, the text removal output of PSSTRNet in each iteration $i_{th}$ is denoted as $I_{out}^i$ and text segmentation result as $M^i$.

\textbf {Region content Loss.} We use $L_1$ loss as the region content loss for text and non-text region reconstruction:

\begin{equation}
\begin{split}
    L_{rc} = \gamma_1 * \sum_{i}|| M_{gt} \odot (I_{out}^i - I_{gt})||_1+\\
    \gamma_2 * \sum_{i}||(1 - M_{gt})\odot (I_{out}^i - I_{gt})||_1.
    \label{eq5}
\end{split}
\end{equation}
where $\gamma_1 \gamma_2$ is set to be 50,10.

\textbf {Perceptual Loss.} We employ the perceptual loss\cite{johnson2016perceptual} in Eq.\eqref{eq11}. $\Phi_i$ is the activation map of the $i$-th layer of the VGG-16 backbone. $H_n$, $W_n$, and $C_n$ denotes the height, width and channel numbers of feature maps outputted from $n_{th}$ layer of VGG-16.
\begin{equation}
\begin{split}
	L_{p} = \sum_{i}\sum_{n}\frac{1}{H_nW_nC_n}||\Phi_i(I_{out}^i) - \Phi_i(I_{gt})||_1
\end{split}
\label{eq11}
\end{equation}

\textbf {Style Loss.} We compute the style loss \cite{2016Image} as Eq.\eqref{eq13}, where $G_n$ is the Gram matrix constructed from the selected activation maps.
\begin{equation}
\begin{split}
	L_{s} =  \sum_{i}\sum_{n}\frac{1}{H_nW_nC_n}||G_n(I_{out}^i)_T\cdot \cr G_n(I_{out}^i) - G_n(I_{gt})^T\cdot G_n(I_{gt})||_1
\end{split}
\label{eq13}
\end{equation}

\textbf {Segmentation Loss.} For learning of text segmentation module, $L_{seg}$ in Eq.\ref{eq14} is formulated as dice loss \cite{milletari2016fully}.

\begin{equation}
	L_{seg} = \sum_{i}{1-\frac{2\sum_{x,y}(M^i(x,y)\times M_{gt}(x,y))}{\sum_{x,y}(M^i(x,y)^2 \times M_{gt}(x,y))}}*\gamma_i
\label{eq14}
\end{equation}
where $\gamma_i$ is set to be 1,2,3. ($x$, $y$) denotes each pixel coordinate in the image. 

In Summary, the total loss for training PSSTRNet is the weighted combination of all the above loss functions.
\begin{equation}
\begin{split}
	L_{total} =  200*L_{s} + 0.1 * L_{p} + L_{rc} + L_{seg}
\end{split}
\end{equation}

\section{Experiments and Results}

\subsection{Datasets and Evaluation Metrics}
\textbf{SCUT-Syn}. This synthetic dataset only includes English text instances, including 8,000 images for training and 800 images for testing. More details can be found in \cite{gupta2016synthetic}.

\textbf{SCUT-EnsText}. It contains 2,749 training images and 813 test images which are collected in real scenes. More descriptions refer to \cite{2020EraseNet}.

\textbf{Evaluation Metrics:} For detecting text on the output images, we employ a text detector CRAFT\cite{2019Character} to calculate recall and F-score. The lower, the better. Six alternative metrics are adopted for measurement the equality of the output images, i.e, PSNR, MSE, MSSIM, AGE, pEPs, and pCEPS\cite{2018EnsNet}. A higher MSSIM and PSNR, and a lower AGE, pEPs, pCEPS, and MSE indicate better results.

\subsection{Implementation Details}

We train PSSTRNet on the training set of SCUT-EnsText and SCUT-Syn and evaluate them on their testing sets, respectively. The masks are generated by subtraction from the input images and the labels. We use dilated process for covering as much of the text area as possible. We follow \cite{2020EraseNet} to apply data augmentation during the training stage. The model is optimized by adam optimizer. Experimentally, we set the iteration number to be 3. The learning rate is set to be 0.001. The learning rate decayed by 50\% every 10 epochs. The PSSTRNet is trained by a single NVIDIA GPU with a batch size of 6 and input image size of 256$\times$256.

\begin{table}[hb]
\center
  \caption{Ablation study results of different modules effect on SCUT-Text.}
  \resizebox{\linewidth}{!}{
  \begin{tabular}{|cc|c|c|c|c|c|c|}
    \hline
    \multicolumn{2}{|c|}{Iterations}&PSNR&MSSIM&MSE&AGE&pEPs&pCEPs\\
    \hline
     \multicolumn{2}{|c|}{1It.} & 32.97 & 96.41 & 0.0017 & 2.0742 & 0.0180 & 0.0105 \\
    \hline
    \multicolumn{2}{|c|}{2It.} & 34.09 & 96.40 & \textbf{0.0014} & 1.7788 & 0.0144 & 0.0077 \\
    \hline
    \multicolumn{2}{|c|}{3It.} & 32.44 & 95.56 & 0.0028 & 2.4506 & 0.0209 & 0.0125 \\
    \hline
    \multicolumn{2}{|c|}{4It.} & 32.15 & 95.69 & 0.0020 & 2.1221 & 0.0184 & 0.0100 \\
    \hline
    \multicolumn{2}{|c|}{2It.+AF} & 34.13 & 96.42 & \textbf{0.0014} & 1.7388 & 0.0142 & 0.0075 \\
    \hline
    \multicolumn{2}{|c|}{3It.+AF} & \textbf{34.65} & \textbf{96.75} & \textbf{0.0014} & \textbf{1.7161} & \textbf{0.0135} & \textbf{0.0074} \\
    \hline
    \multicolumn{2}{|c|}{4It.+AF} & 33.02 & 96.46 & 0.0017 & 2.0084 & 0.0177 & 0.0098 \\
    \hline
\end{tabular}}
\label{tab1}
\end{table}

\subsection{Ablation Study}

In this section, we study the effect of the number of iterations and the adaptive fusion method on the SCUT-Text dataset. In total, we conduct seven experiments by designing the network with 1) one iteration (1It.), 2) two iterations (2It.), 3) three iterations (3It.), 4) four iterations (4It.), 5) two iterations with adaptive fusion(2It.+AF), 6) three iterations with adaptive fusion(3It.+AF), 7) four iterations with adaptive fusion(4It.+AF). All experiments use the same training and test settings.

Qualitative and quantitative results are illustrated in Fig.\ref{Ablation} and table\ref{tab1}, respectively. We can see that the network generates the best STR results with two iterations if only considering iteration times (i.e., comparing results in the first four experiments). This arises from that the information is lost in increasing iterations using encoder-decoder architecture. By adding an adaptive fusion strategy, the model with three iterations (3It.+AF) gets the best results. It is because adaptive fusion utilizes previous removal results and could also get more erased regions on text when increasing iterations. As shown in (b)(c)(d) of Fig.\ref{Diff_stage}, our method gets a roughly segmentation result at $1_{st}$ iteration and extracts the rest part of the text segments, and cleaner text removal results in the following iterations. However, We find that the style of intermediate result is distorted when increasing the iterative times to 4 or larger. The decreasing qualitative results of 7$_{th}$ experiment 4It.+AF in table\ref{tab1} reflect this point.

\subsection{Comparison with State-of-the-Art Approaches}

\begin{table*}
\center
  \caption{Comparison with SOTA methods and proposed method on SCUT-EnsText. R: Recall; P: Precision; F: F-score.}
  \begin{tabular}{|c|c|c|c|c|c|c|c|c|c|c|}
    \hline
    \multirow{2}{*}{Method}&\multicolumn{6}{c|}{Image-Eval}&\multicolumn{3}{c|}{Detection-Eval(\%)}\\\cline{2-10}
    &PSNR $\uparrow$ &MSSIM$\uparrow$ &MSE$\downarrow$ &AGE$\downarrow$ &pEPs$\downarrow$ &pCEPs$\downarrow$ &P$\downarrow$ &R$\downarrow$ &F$\downarrow$\\
    \hline
    Original Images  & - & - & - & - & - & - & 79.8 & 69.7 & 74.4 \\
    \hline
    Pix2pix & 26.75 & 88.93 & 0.0033 & 5.842 & 0.048 & 0.0172 & 71.3 & 36.5 & 48.3 \\
    \hline
    Scene Text Eraser & 20.60 & 84.11 & 0.0233 & 14.4795 & 0.1304 & 0.0868 & 52.3 & 14.1 & 22.2 \\
    \hline
    EnsNet & 29.54 & 92.74 & 0.0024 & 4.1600 & 0.2121 & 0.0544 & 68.7 & 32.8 & 44.4 \\
    \hline
    EraseNet& 32.30 & 95.42 & 0.0015 & 3.0174 &0.0160 & 0.0090 & 53.2 & 4.6 & 8.5 \\
    \hline
    PERT(official) & 33.25 & \textbf{96.95} & \textbf{0.0014} & 2.1833 & 0.0136 & 0.0088 & 52.7 & \textbf{2.9} & \textbf{5.4} \\
    \hline
    PSSTRNet(Ours) & \textbf{34.65} & 96.75 & \textbf{0.0014} & \textbf{1.7161} & \textbf{0.0135} & \textbf{0.0074} & \textbf{47.7} & 5.1 & 9.3 \\
    \hline
\end{tabular}
\label{tab2}
\end{table*}

\begin{table*}
\center
  \caption{Comparison with SOTA methods and proposed method on SCUT-Syn.}
  \begin{tabular}{|c|c|c|c|c|c|c|c|c|c|}
    \hline
    Method &PSNR $\uparrow$ &MSSIM$\uparrow$ &MSE$\downarrow$ &AGE$\downarrow$ &pEPs$\downarrow$ &pCEPs$\downarrow$&Parameters$\downarrow$&Inference Time$\downarrow$\\
    \hline
    Pix2pix & 25.16 & 87.63 & 0.0038 & 6.8725 & 0.0664 & 0.0300 & 54.4M & \textbf{2.96ms}  \\
    \hline
    Scene Text Eraser & 24.02 & 89.49 & 0.0123 & 10.0018 & 0.0728 & 0.0464 & 89.16M & 18.45ms  \\
    \hline
    EnsNet & 37.36 & 96.44 & 0.0021 & 1.73 & 0.0276 & 0.0080 & 12.4M & 5.1 ms \\
    \hline
    EraseNet & 38.32 & 97.67 & \textbf{0.0002} & 1.5982 & 0.0048 & \textbf{0.0004}& 19.74M & 8.67ms \\
    \hline
    PERT(official) & \textbf{39.40} & 97.87 & \textbf{0.0002} & 1.4149 & 0.0045 & 0.0006 & 14.00M & - \\
    \hline
    PSSTRNet(Ours) & 39.25 & \textbf{98.15} & \textbf{0.0002} & \textbf{1.2035} & \textbf{0.0043} & 0.0008 & \textbf{4.88M} & 14.9ms \\
    \hline
    
\end{tabular}
\label{tab3}
\end{table*}

We compare our proposed PSSTRNet with five state-of-the-art methods: Pix2pix\cite{2017Image}, STE\cite{2017Scene}, EnsNet\cite{2018EnsNet}, EraseNet\cite{2020EraseNet} and PERT \cite{wang2021pert}, on both SCUT-EnsText and SCUT-Syn datasets. We retrain all the models with the setting as official reported, but input the image of size 256$\times$256. The source code of PERT is not released currently, so we do not show its qualitative results.

\textbf{Qualitative Comparison}. As shown in the 1st row of Fig.\ref{SOTA}, our model can preserve more information in non-text areas while erasing text regions cleaner. Compared with other state-of-the-art methods, the results of our proposed PSSTRNet have significantly fewer color discrepancies and blurriness, especially in 1st, 2nd, and 4th lines. It demonstrates our model could generate more semantically elegant results on text removal and background inpainting results.

\textbf{Quantitative Comparison}. As shown in Table \ref{tab2} and \ref{tab3}, our method produces the best scores on most text removal evaluation protocols for both SCUT-EnsText and SCUT-Syn datasets. Furthermore, our model has the minimum number of parameters, which only has about one-third of the parameters of PERT that also implements STR in a progressive way.

\section{limitation}

As shown in the 3rd row of Fig.\ref{SOTA} and Fig.\ref{Ablation}, there are still some texts that are not be removed. Besides, our model's inference time is longer than others since we apply iterative processes. Hence, there is still some improvement space of our method in terms of text detection and inference time. Combining our work with a better scene text detector may lead to better results.

\section{Conclusion}
In this paper, we present a light-weighted progressive network PSSTRNet for scene text removal in images. It is based on an encoder-decoder structure with a shared encoder and two decoder branches for progressive text segmentation and text removal respectively. A Mask Updated module is developed to gradually acquire more and more complete and accurate text masks for better guidance. Instead of using the output from the final iteration, we aggregate the results in each iteration by adaptive fusion. Experimental results indicate that the proposed method achieves state-of-the-art performance on both synthetic and real-world datasets while maintaining low complexity.

\bibliographystyle{IEEEbib}
\bibliography{icme2022template}

\end{document}